\documentclass[a4paper]{styles/svproc}
\usepackage{math}
\usepackage{amsmath}
\usepackage{flushend}
\usepackage{tabularx} 
\usepackage{url}

\newcommand{\orange}[1]{\textcolor{orange}{#1}}

\usepackage{tcolorbox}
\definecolor{mycolor}{rgb}{0.811, 0.882, 0.952}
\newtcolorbox{mybox}{colback=gray!10!white,colframe=black!50!black, boxrule=0.4pt,arc=6pt, left=6pt,right=6pt,top=5pt,bottom=5pt}

\begin{document}
\mainmatter              % start of a contribution
%
% \title{MOSU: Global Navigation with Multiple-\\
% modality Input and Scene Understanding}

\title{MOSU: Autonomous Long-range Robot Navigation with Multi-modal Scene Understanding}
% and On-road Scene Understanding}

%
\titlerunning{MOSU: Autonomous Long-range Robot Navigation}  % abbreviated title (for running head)
\author{Jing Liang$^\dagger$, Kasun Weerakoon$^\dagger$, Daeun Song$^\ddagger$, Senthurbavan Kirubaharan$^\ddagger$, Xuesu Xiao$^\ddagger$, \and Dinesh Manocha$^\dagger$}
\authorrunning{Jing Liang et al.} % abbreviated author list (for running head)
\tocauthor{Jing Liang, Kasun Weerakoon, Daeun Song, , Xuesu Xiao and Dinesh Manocha}
\institute{$^\dagger$University of Maryland, College Park MD, 20740, USA \\
$^\ddagger$Goerge Mason University, Fairfax, VA, 22030, USA
}

\maketitle              % typeset the title of the contribution

\begin{abstract}
% \vspace{-1em}
We present MOSU, a novel autonomous long-range navigation system that enhances global navigation for mobile robots through multimodal perception and on-road scene understanding. 
MOSU addresses the outdoor robot navigation challenge by integrating geometric, semantic, and contextual information to ensure comprehensive scene understanding. The system combines GPS and QGIS map-based routing for high-level global path planning and multi-modal trajectory generation for local navigation refinement. For trajectory generation, MOSU leverages multi-modalities: LiDAR-based geometric data for precise obstacle avoidance, image-based semantic segmentation for traversability assessment, and Vision-Language Models (VLMs) to capture social context and enable the robot to adhere to social norms in complex environments. This multi-modal integration improves scene understanding and enhances traversability, allowing the robot to adapt to diverse outdoor conditions. 
We evaluate our system in real-world on-road environments and benchmark it on the GND dataset, achieving a 10\% improvement in traversability on navigable terrains while maintaining a comparable navigation distance to existing global navigation methods.


% \vspace{-0.5em}
\end{abstract}

\section{Introduction}
% \vspace{-0.5em}
% Motivation, Problem Statement, Related Work (one page)

Global navigation has witnessed significant advancements in recent years \cite{ gao2019global}, playing a crucial role in applications such as autonomous driving \cite{ yurtsever2020survey}, logistics \cite{hoffmann2018regulatory}, and search and rescue \cite{davids2002urban}. However, several key challenges remain. Many existing approaches depend on highly accurate global maps and precise localization \cite{ozturk2022review}, which are costly and difficult to maintain at scale. Additionally, scene understanding, particularly traversability analysis and socially compliant navigation, poses a significant challenge for generating safe and context-appropriate trajectories. Ensuring reliable long-range navigation across diverse and dynamic real-world environments further complicates the problem, as the system must adapt to varying terrain, obstacles, and social contexts.
Addressing these challenges requires an approach that integrates global planning with multimodal perception and adaptive local planning to enable robust and scalable autonomous navigation.

% Global navigation has a significant development in recent years \cite{ganesan2022global, gao2019global, ort2018autonomous}. It plays a crucial role in tasks such as autonomous driving \cite{ort2018autonomous, yurtsever2020survey}, logistics \cite{hoffmann2018regulatory, chen2021adoption}, search and rescue \cite{davids2002urban}, etc. However, global navigation is still challenging: (1) Most of previous approaches rely on highly accurate global maps and localization \cite{gasparetto2015path, ozturk2022review, guo2023efficient}, which are expensive to maintain at large scales; (2) Scene understanding is challenging (e.g., traversability analysis and social navigation) for safe and efficient trajectory generation; (3) the demand for reliability in long-range navigation across diverse real-world conditions; and (4) the need for real-time, onboard computation with low latency and adaptability. To address these challenges, we propose MOSU, a novel global navigation system.

Maintaining an accurate map is time- and labor-intensive, and various temporal conditions can degrade planning performance~\cite{gasparetto2015path, ozturk2022review}. However, humans do not require highly accurate maps for global navigation. Given Google Maps, we can complete most long-range navigation tasks, such as commuting from home to work, traveling in a new city, or walking on trails. Inspired by this observation, instead of maintaining a highly sophisticated global map, we propose separating this task into two easily accessible and generalizable components: routing and trajectory generation. Routing provides raw latitude and longitude directions, while trajectory generation determines the traversability of the environment to guide the robot to the next GPS location.

Scene understanding is essential for trajectory generation in outdoor robot navigation. It is a complex process that requires the integration of geometric perception \cite{kong2023robo3d}, semantic comprehension \cite{kim2024learning, 9699042}, and contextual awareness \cite{sathyamoorthy2024convoi}. Geometric perception enables obstacle avoidance, semantic understanding distinguishes traversable paths, and contextual awareness ensures socially compliant behavior. Foundation models have recently demonstrated strong capabilities in capturing social contexts~\cite{song2024tgs, song2024vlm}. However, many trajectory generation approaches prioritize only one or two aspects due to the high computational costs on resource-constrained onboard systems \cite{liang2024mtg, liang2024dtg}, leading to incomplete environmental understanding and limiting navigation reliability.
To address this, our approach incorporates multiple modalities, including LiDAR-derived geometric confidence~\cite{liang2024mtg}, image-based color semantics~\cite{cheng2021mask2former}, VLM-based context awareness~\cite{song2024tgs}, and robot odometry that achieves comprehensive on-road scene understanding for robot navigation. 

\textbf{Problem Statement: } 
Designing a long-range navigation system with autonomous routing and trajectory generation by leveraging multimodal perception and VLMs to achieve comprehensive scene understanding, enhancing both traversability and social awareness. 
%Designing a long-range navigation system with autonomous routing and trajectory generation capabilities by leveraging multimodal perception and VLMs for traversability estimation and social cues. 

% \todo{also contain keywords: long-range, campus environment.}

% \vspace{-0.5em}
\section{Our Approach}
% Technical Approach (one page)
% \vspace{-0.5em}

\begin{figure}
% \vspace{-2em}
    \centering
    \includegraphics[width=\linewidth]{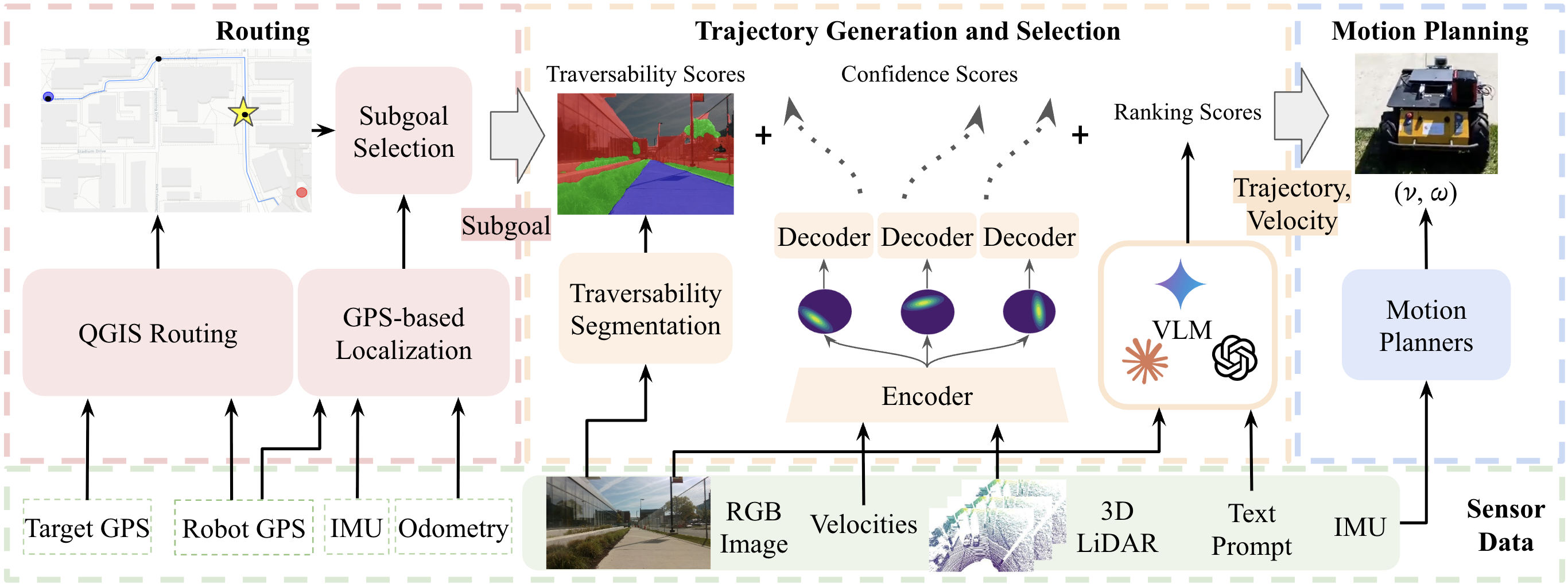}
    \caption{\small{Overall System Architecture. MOSU leverages QGIS and GPS to generate long-range waypoints, serving as high-level guidance for its trajectory generation system. The trajectory generation system provides local trajectories by integrating multimodal perception cues for traversability assessment and leveraging social cues from Vision-Language Models (VLMs) to ensure social compliance in navigation. This hierarchical approach enables robust, adaptive, and context-aware trajectory generation for long-range autonomous navigation.} }
    \label{fig:architecture}
%\vspace{-2em}
\end{figure}

We propose a novel global navigation system, MOSU, with \textbf{M}ulti-modal perception and \textbf{O}n-road \textbf{S}cene \textbf{U}nderstanding for mobile robots. While traditional outdoor navigation systems rely on detailed global maps and integrate global path planning with local motion planning, MOSU instead decomposes global path planning into two separate components: routing and trajectory generation. Specifically, we leverage QGIS and GPS data for high-level routing and use multi-modal sensor inputs for low-level trajectory generation and scene understanding. As shown in Fig.~\ref{fig:architecture}, the system consists of three stages: routing, trajectory generation, and motion planning.

\subsection{Routing} 
The routing stage computes a high-level path from the robot's current GPS location to the target by generating a sequence of intermediate GPS subgoals. 
These subgoals are computed by public satellite routing service, such as Google or Openroute. % we are not going to mention a map, because we are not uisng accurate map in the navigation.
Due to the limited precision of GPS sensors (5$\mathrm{m}$)\cite{van2015world}, the sub-goals can only serve as directional guides towards the target rather than exact positions, and if the sub-goals are too close the noise will lead the robot to move into non-traversable areas. To prevent this short-range effect of the GPS sub-goals, we space the sub-goals approximately 50$\mathrm{m}$ apart, corresponding to the robot's perceptual range with 3D LiDAR and RGB sensors. 
As the robot navigates, we convert the current GPS sub-goal into the robot's coordinates and continuously monitor its approximate distance to the current sub-goal and updates to the next one upon reaching a predefined proximity threshold, 10$\mathrm{m}$. This strategy enables scalable long-range navigation while allowing the local trajectory generation module to handle fine-grained motion decisions based on real-time sensor observations.

\subsection{Trajectory Generation}

Given the current GPS sub-goal, the robot needs a trajectory to navigate toward it. In complex outdoor environments, the robot must handle obstacles, varying terrain traversability, social norms, and traffic rules. We decompose this task into two subproblems: traversability analysis and social navigation. During trajectory generation, we apply different methods to address these challenges and generate trajectories that are both traversable and socially compliant, using multi-modal sensor inputs such as RGB images, robot velocities, 3D LiDAR point clouds, and text prompts (Fig.~\ref{fig:architecture}).

For traversability analysis, relying solely on geometric or color information is insufficient to fully understand the environment’s traversability (i.e., which areas are safe for the robot to traverse), as shown in Fig. \ref{fig:challenge}. Our system integrates both geometric and color information for more accurate analysis. Social navigation is highly intuitive and depends on the common sense and the culture of the country. Therefore, we utilize generalized vision-language models (VLMs) to address the social navigation problem.

Our trajectory generation method consists of four key components: (1) CVAE-based multiple trajectories generation, which models the distribution of feasible trajectories based on LiDAR and velocity inputs, (2) semantic segmentation for traversability analysis, which leverages RGB images to identify navigable surfaces (3) %surface normal estimation, which captures fine-grained geometric cues for local terrain understanding, (4) 
VLM-based trajectory ranking, which ranks candidate trajectories based on social compliance using overlaid image and natural language prompts.

%Different from the traditional point cloud segmentation methods for traversability analysis, 
\textbf{CVAE-based Multiple Trajectories Generation:} This model is to understand the geometric information of the environment for traversability analysis through 3D Lidar point clouds. However, directly modeling the environment using point cloud is memory costly and the segmentation of a large point cloud is also heavy~\cite{guo2020deep, sohail2024advancing}. To address the issue, we adopt MTG~\cite{liang2024mtg}, which is a very light-weighted learning-based approach to generate multiple candidate trajectories to cover traversable regions in front of the robot. MTG employs a Conditional Variational Autoencoder (CVAE)~\cite{cvae} to model the distribution of feasible trajectories. 
Given sensor input $\x=\set{\l, \v}$, where $\l\in \mathcal{L}$ represents a sequence of LiDAR observations and $\v\in\mathcal{V}$ denotes the robot's historical velocities, the model generates a set of $N$ candidate trajectories $\mathcal{T}=\{\tau_1,...,\tau_N\}$. For each trajectory $\tau_n\in\ct$, we have the following formulation: %%% HERE
% \vspace{-0.5em} 
\begin{align}
     % p(\tau|\x) &\approx \frac{1}{S}\sum_{s=1}^S p(\tau|\z^{(s)},\c), \;\;\;\z^{(s)} \sim \cn(\mu, \nu),
     &p_\theta(\tau_n|\x) = p_\theta(\tau_n | \z_n, \c), \;\;\;\z_n \sim \cn(\mu_n, \nu_n), \\
     &\c = f_\theta(\x), \;\;\; \mu, \nu = g_\theta(\x)
     \vspace{-3em}, \;\;\; \cn(\mu_n, \nu_n) = h_\theta(\cn(\mu, \nu), \c)
     \label{eq:trajectory_generation}
\end{align}
% \vspace{-0.5em}
We slightly abuse the notation of $\theta$ to represent the parameters of all neural networks. $\c$ is the conditional vector obtained from an encoder network $f_\theta(\cdot)$. $g_\theta(\cdot)$ is the latent encoder network that takes the sensor inputs $\x$ and predicts the parameters, mean $\mu$ and variance $\nu$, of a Gaussian distribution $\cn(\mu, \nu)$. In our approach, the LiDAR data is processed by PointCNN~\cite{li2018pointcnn}, and the velocities are processed by a sequence of linear layers, as shown in~\cite{liang2024mtg}. Then we linearly transform the distribution to $N$ Gaussian distributions $\cn(\mu_n, \nu_n)$ by the learnable linear transformation (neural network) $h_\theta(\cdot)$. We then sample the latent vector $\z_n$ from the distribution $\cn(\mu_n, \nu_n)$ and use it as the input of the trajectory decoder $p_\theta(\cdot)$ to generate trajectory $\tau_n$. The predicted variance $\nu$ represents the uncertainty in the latent space and is used to compute a confidence score $c_\tau$ for each trajectory. This formulation takes the LiDAR geometric information as input and generates trajectories to cover geometrically traversable areas in front of the robot.

\textbf{Semantic Segmentation for Traversability Analysis:} 
However, this model struggles to accurately detect the boundaries between adjacent traversable and non-traversable areas when they have similar geometric structures (e.g., off-road mud vs. sidewalks), as shown in Fig.\ref{fig:challenge} (c). To address this limitation, we incorporate semantic information by using Mask2Former\cite{cheng2021mask2former}, an RGB-based semantic segmentation model. The model segments the image into multiple regions with different semantic categories by predicting the class label of each pixel. We define five traversability categories~\cite{gnd}: road, sidewalk, vegetation, building, and others. Different types of robots have different traversable areas; for wheeled robots, we constrain them to operate only within the sidewalk and road traversability categories.

Given the segmented image, we overlay the set of candidate trajectories $\mathcal{T}$, generated by the CVAE-based trajectory generator, onto the image and evaluate their traversability scores. First, we use the Bresenham algorithm~\cite{bresenham1998algorithm} to convert trajectory waypoints into connected pixels in the image. Then, the semantic traversability score $t_\tau$ is computed for each trajectory as the ratio of pixels falling in traversable areas to the total number of pixels along the trajectory.

% \textbf{Surface normal Estimation:} 
% \todo{Jing}

\textbf{VLM-based Trajectory Ranking:}
To ensure social compliance, we incorporate VLMs~\cite{song2024tgs, song2024vlm} to understand social cues from the robot's observation. As in VL-TGS~\cite{song2024tgs}, we project the trajectories onto the image and, from right to left, we assign numbers to the trajectories in sequence, according to the pixel positions of the last waypoint of each trajectory. Then, VLMs take the overlayed RGB image and a text prompt as input. % and output the ranking score, $r_\tau= \frac{1}{N}(N - p_n)$, of the trajectories with chain-of-thought logic to satisfy the criteria in the text prompt, where $p_n$ is the ranked position of the trajectory $\tau_n$. 
The following is the prompt input to the model:
\begin{mybox}
The \orange{N} trajectories are labeled with numbers \orange{[0-N-1]} from right to left in sequence. The goal is \orange{K} meters at \orange{[Right Front]}. Rank trajectories for social navigation. 

\begin{enumerate}
\item keep away from the groups of pedestrians. The robot has three mode, Normal, Slow, and Stop. If the people are approaching, the robot needs to Slow. If people are too close or there is no open space, the robots Stops.

\item follow the traffic rules, and if going across the street, the robot should keep in crosswalks.

\item recognize the traffic signs and behave accordingly. 

\item avoid off-road terrain for small wheeled robots. 

\end{enumerate}

Given the picture, the target is at \orange{K meters Front Left}. Rank the trajectories by the criteria. output the format: [robot mode], [ranked numbers], reason
\end{mybox} 
\noindent where the orange text indicates variables based on the current sub-goal and candidate trajectories. The model output consists of three parts: (1) the robot’s current velocity mode—either slow or normal; (2) the ranking of trajectories based on social and traffic compliance; and (3) the reasoning behind the ranking, enabling chain-of-thought understanding of the environment. From the ranking of trajectories, we calculate the ranking score, $r_\tau= \frac{1}{N}(N - p_n)$, where $p_n$ is the ranking of the trajectory $\tau_n$. Leveraging chain-of-thought reasoning guided by the prompt, the VLM selects trajectories that are both socially compliant and contextually appropriate.
%, of the trajectories with chain-of-thought logic to satisfy the criteria in the text prompt, where $p_n$ is the ranked position of the trajectory $\tau_n$.

While each component contributes complementary information, they also come with individual limitations in scene understanding. For example, RGB-based segmentation lacks geometric precision and often fails to capture elevation changes such as curbs. Learning-based models may also misinterpret out-of-distribution regions under unfamiliar conditions, as illustrated in Fig.~\ref{fig:challenge} (d). To mitigate these limitations, we aggregate the outputs from all components to compute a final score for each candidate trajectory. Since trajectories are generated in real time, we further incorporate multiple consecutive frames, transforming them into the current robot frame to ensure consistency. The optimal trajectory $\tau$ is selected by maximizing a weighted sum of scores:
% \vspace{-0.5em}
% \begin{align}
%     \tau = \max\set{\beta_1 c_ \tau ^i + \beta_2 
%  t_\tau ^i + \beta_3 p_\tau^i + \beta_4 g_\tau^i},
%  \label{eq:trajectory}
% \end{align}
\begin{align}
    i^* = \arg\max_{i \in [1, N]} \left\{ \beta_1 c_\tau^i + \beta_2 t_\tau^i + \beta_3 r_\tau^i + \beta_4 g_\tau^i \right\}, \;\;\;
    \tau = \tau^{i^*},
    \label{eq:trajectory}
\end{align}
where $\beta_{1,2,3,4}$ are weights of the components, and $i\in[1,N]$ represents the index of the $N$ generated trajectories. $c_\tau^i$ denotes the confidence score from geometric information, $t_\tau^i$ represents a semantic traversability score, and $r_\tau^i$ corresponds to the ranking score from VLMs, where higher-ranked trajectories receive greater weights. $g_\tau^i$ is the distance score to the nearest GPS subgoal, the closer, the higher.

\subsection{Motion Planning} 
This stage generates executable robot actions to follow the selected trajectory $\tau$ from Equation~\ref{eq:trajectory}. We integrate the Dynamic Window Approach (DWA)~\cite{fox1997dynamic}, a widely used reactive local planner that generates safe and feasible motion commands in real time. In addition to standard trajectory following, we apply the velocity mode (normal or slow) predicted by the VLM to constrain the robot’s maximum velocity to social-compliantly move the robot. The normal mode allows a maximum velocity of $1,\mathrm{m/s}$, while the slow mode limits the velocity to below $0.5,\mathrm{m/s}$.

\section{Experiments}
% Experiments completed or scheduled (one page)
The experiment is designed to evaluate the efficacy of the system in two aspects: (1) Trajectory generation, focusing on traversability analysis and social understanding. (2) Overall system performance in real-world long-range navigation compared with other approaches. Experiments are conducted on a computer equipped with an Intel i9 CPU (31 GB RAM) and an Nvidia RTX 3060 GPU (6.4 GB VRAM). We compare our approach against MTG~\cite{liang2024mtg}, DTG~\cite{liang2024dtg}, and VL-TGS~\cite{song2024tgs}, NoMaD~\cite{sridhar2024nomad}, ViNT~\cite{shah2023vint}, and PIVOT~\cite{nasiriany2024pivot} with the dataset. We use Gemini~\cite{GeminiTeam2024} for VLM. 

\begin{figure}
  \centering
  \begin{tabular}{ c c c c}
  \begin{minipage}{.24\linewidth}
   \centering
   \includegraphics[width=\linewidth,height=0.8\linewidth]{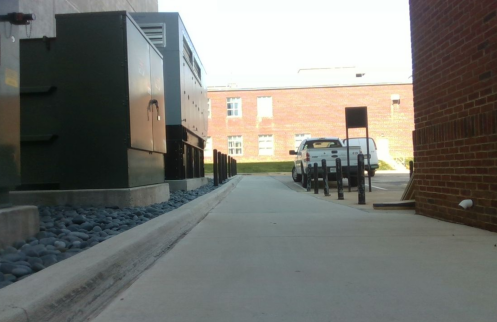} (a) Narrow Space
   \end{minipage}
   &
   \begin{minipage}{.24\linewidth}    \centering
\includegraphics[width=\linewidth,height=0.8\linewidth]{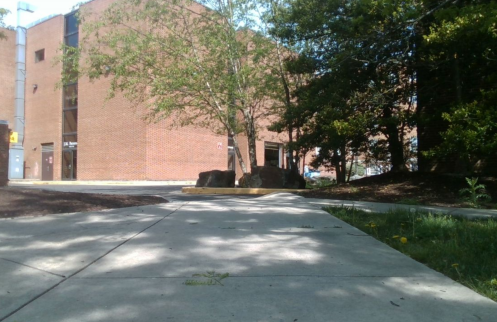}  (b) Off-road Terrain
   \end{minipage} 
   & 
   \begin{minipage}{.24\linewidth}   \centering

   \includegraphics[width=\linewidth,height=0.8\linewidth]{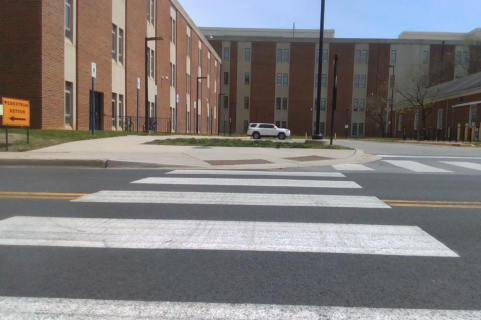} (c) Crosswalks
   \end{minipage}
   & 
   \begin{minipage}{.24\linewidth}    \centering
\includegraphics[width=\linewidth,height=0.8\linewidth]{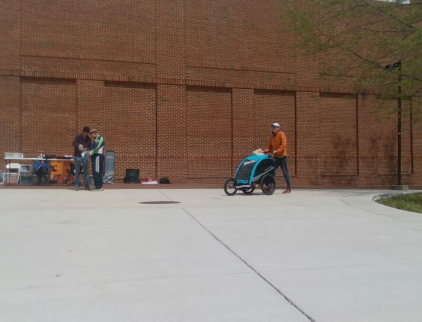} (d) Social Scenarios
\end{minipage}
  \end{tabular}
  % \vspace{-0.5em}
  \caption{\textbf{Complex Outdoor Scenarios:} We evaluate the approach in large-scale, complex outdoor environments with various challenging scenarios, such as (a) narrow spaces, (b) areas with dense off-road vegetation, (c) traffic components (e.g., crosswalks), and (d) social situations involving pedestrians.}
    \label{fig:scenarios}
    % \vspace{-0.5em}
\end{figure}

We evaluate trajectory generation using the GND dataset\cite{gnd}, which covers 10 campuses with diverse scenarios, including urban and rural environments. For traversability analysis and social routine understanding, as shown in the Fig.~\ref{fig:scenarios}, we evaluate the approaches a large-scale dataset (GND) with various challenging scenarios.

The metrics in Table~\ref{tab:gnd_quantitative} are calculated as following:

\textbf{Traversability:} We overlay the generated trajectory onto the traversability map from the GND dataset~\cite{gnd}. The traversability score is then calculated as the percentage of fully traversable waypoints over the entire trajectory length.

\textbf{Distance to Target: } $1 - \frac{d_\tau - d_o}{|\tau|}$, where $d_\tau$ and $d_o$ represent the distances between the target and the current trajectory and between the target and the optimal ground-truth trajectory, respectively. $|\tau|$ denotes the length of the generated trajectory.

% explain the scenario selection for the qualitative evaluation.
% explain the scanrio selection for social understanding

% \textbf{Scheduled Experiments:} We plan to conduct outdoor long-range navigation to qualitatively evaluate the efficacy of global navigation and the reliability of the system during operation. We will show a video to demonstrate the performance in global navigation and pictures to show the generated trajectories across diverse scenarios during this navigation.

\begin{figure}[!ht]

  \centering
  \begin{tabular}{ c c c }
  % \hline
   \begin{minipage}{.32\linewidth}
   \includegraphics[width=\linewidth]{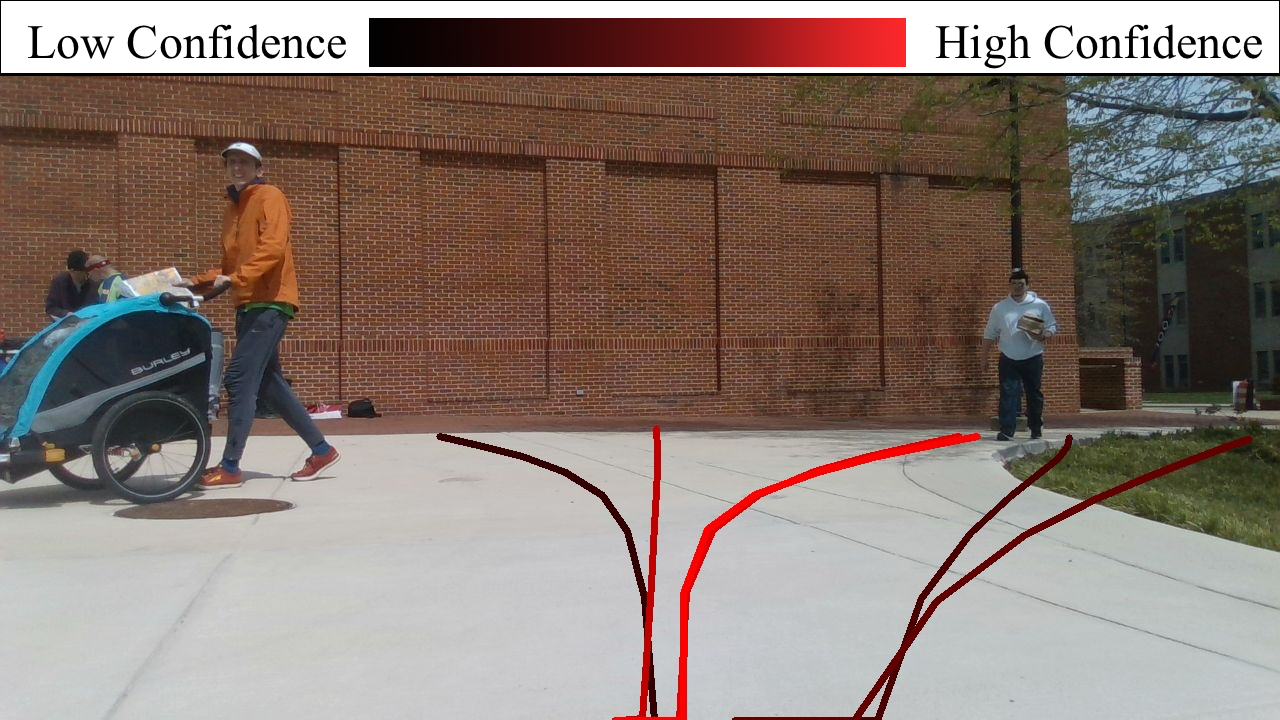}
   \end{minipage}
   & 
   \begin{minipage}{.32\linewidth} \includegraphics[width=\linewidth]{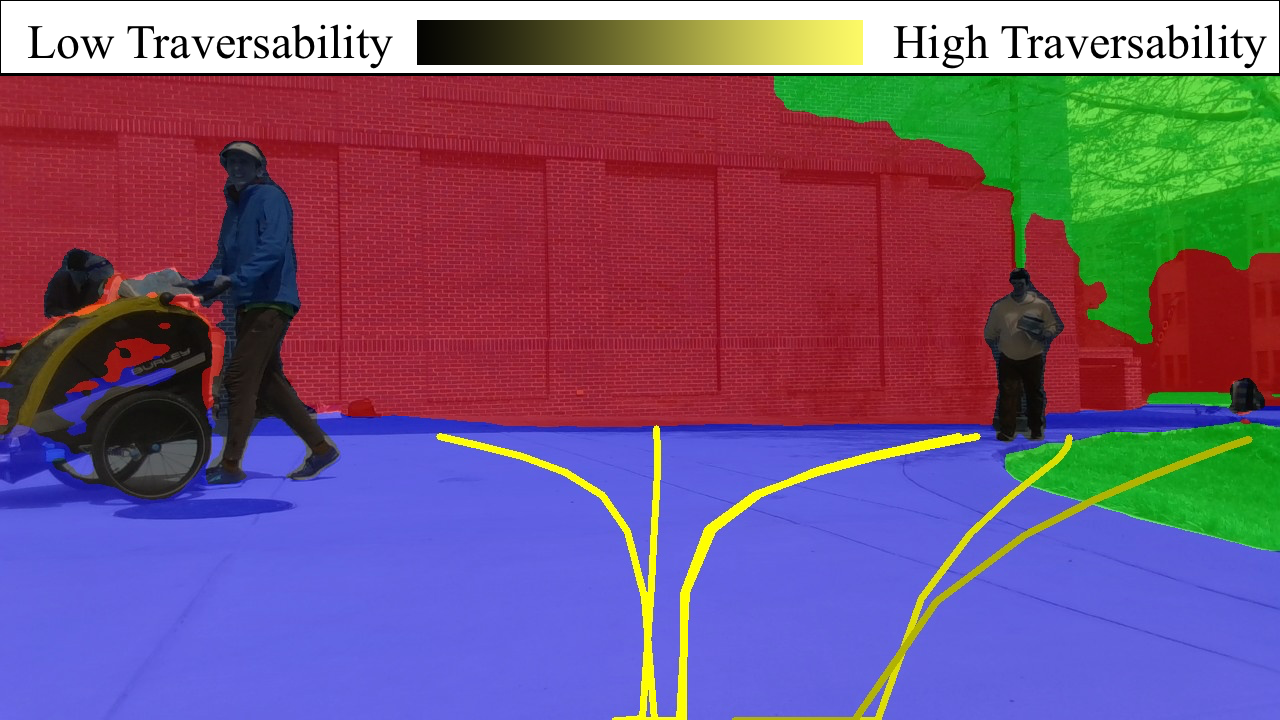} \end{minipage}
   & 
   \begin{minipage}{.32\linewidth} \includegraphics[width=\linewidth]{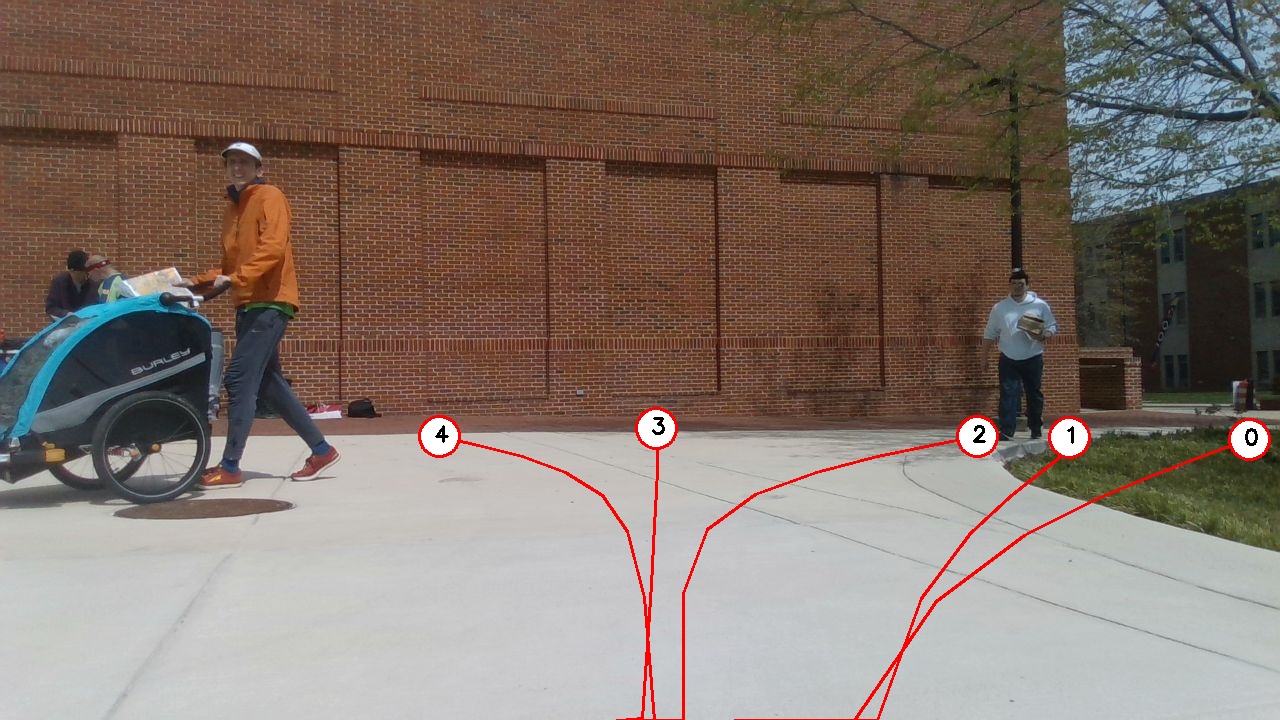} \end{minipage}
   \\ 
   \begin{minipage}{.32\linewidth}
   \includegraphics[width=\linewidth]{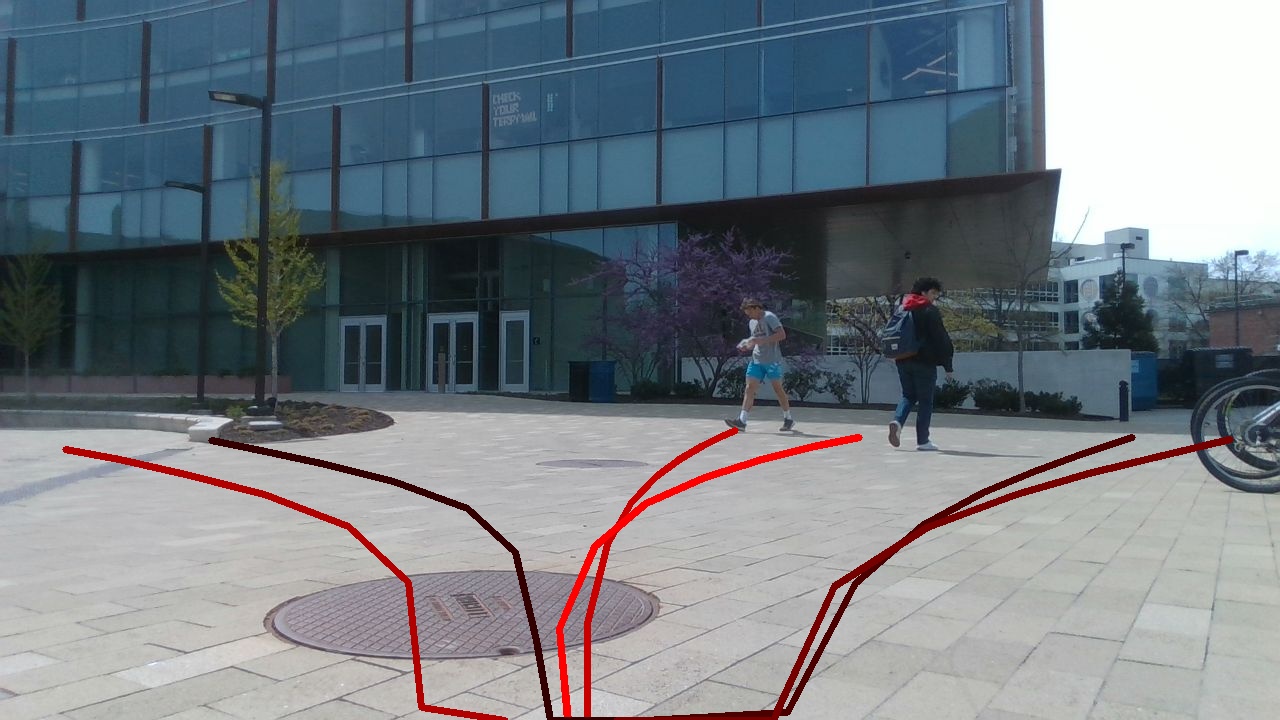}
   \end{minipage}
   & 
   \begin{minipage}{.32\linewidth} \includegraphics[width=\linewidth]{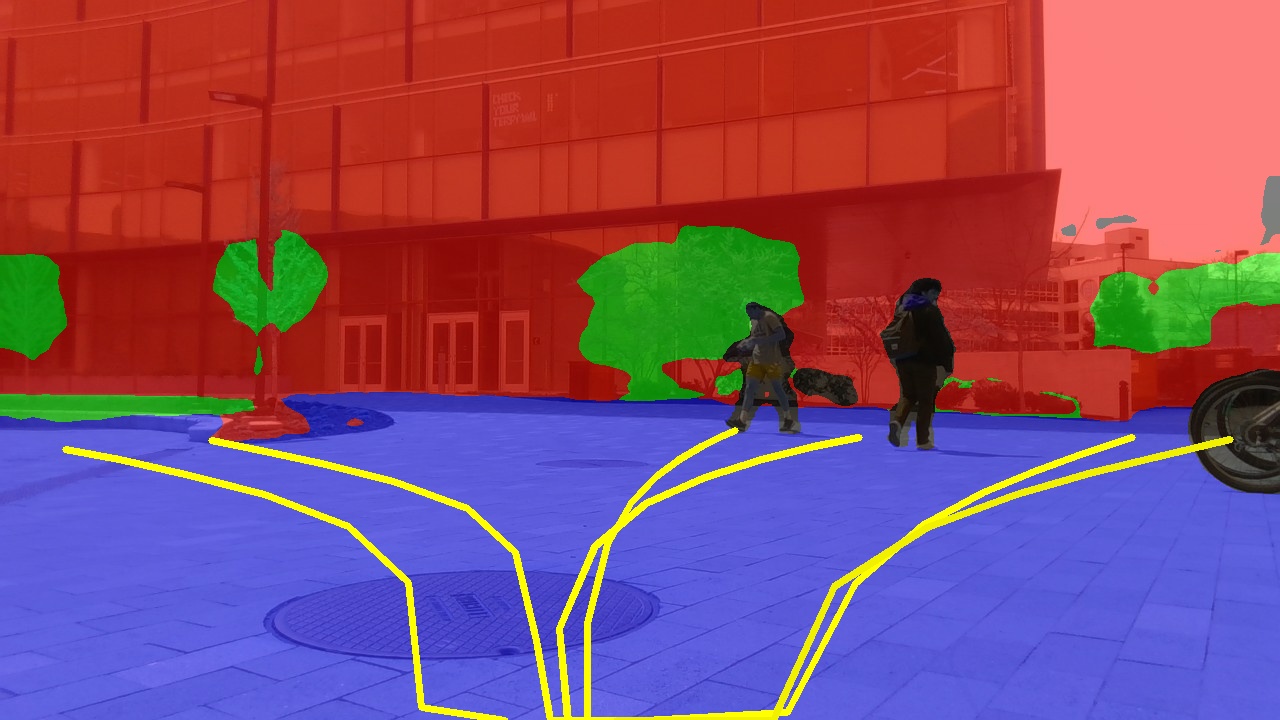} \end{minipage}
   & 
   \begin{minipage}{.32\linewidth} \includegraphics[width=\linewidth]{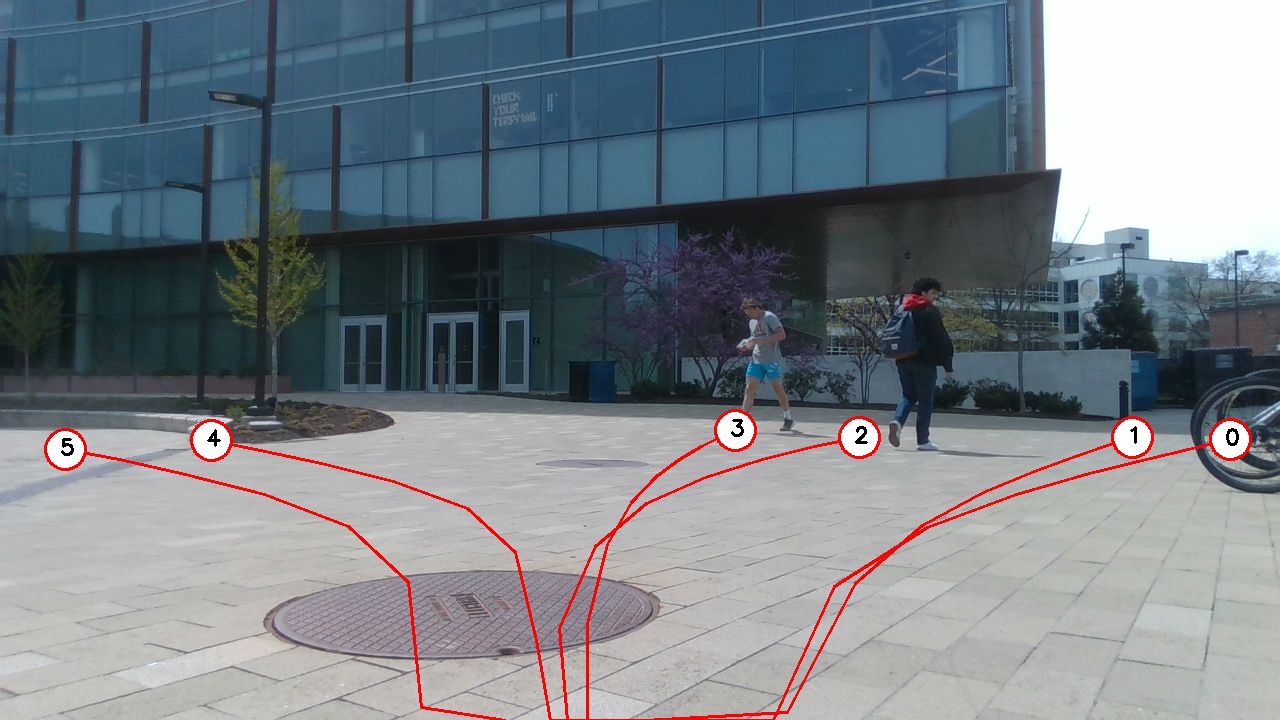} \end{minipage}
   \\ 
   (a) & (b) & (c) \\
   % \footnotesize{Multiple Trajectory Generation} & \footnotesize{Semantic Segmantation} & \footnotesize{VLM-based Trajectory Ranking}
  \end{tabular}
  % \vspace{-0.5em}
  \caption{\small{
  \textbf{Qualitative Evaluation:} %The two rows show how the three components work in the outdoor scenarios. 
  Each column illustrates one of the three components in our trajectory generation pipeline, (a) CVAE-based multiple trajectory generation, (b) semantic segmentation for traversability analysis, and (c) VLM-based trajectory ranking in the outdoor scenarios. In (a) and (b), lighter colors indicate higher scores, confidence scores in (a) and traversability scores in (b). In (c), the VLM ranks trajectories based on social cues. The top example shows a ranking of [3, 0, 1, 2, 4]. The bottom example shows a ranking of [4, 5, 0, 1, 2, 3].}}
  \label{fig:gnd_qualitative}
  % \vspace{-0.5em}
\end{figure}

\begin{figure}
  \centering
  \begin{tabular}{ c c c c}
  \begin{minipage}{.24\linewidth}
   \centering
   \includegraphics[width=\linewidth,height=0.8\linewidth]{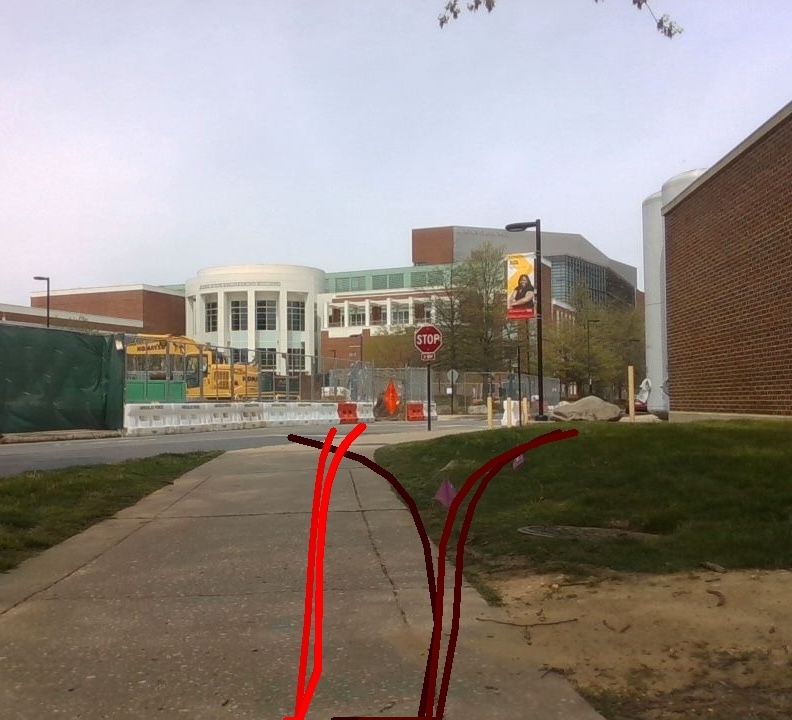} (a)
   \end{minipage}
   &
   \begin{minipage}{.24\linewidth}    \centering
\includegraphics[width=\linewidth,height=0.8\linewidth]{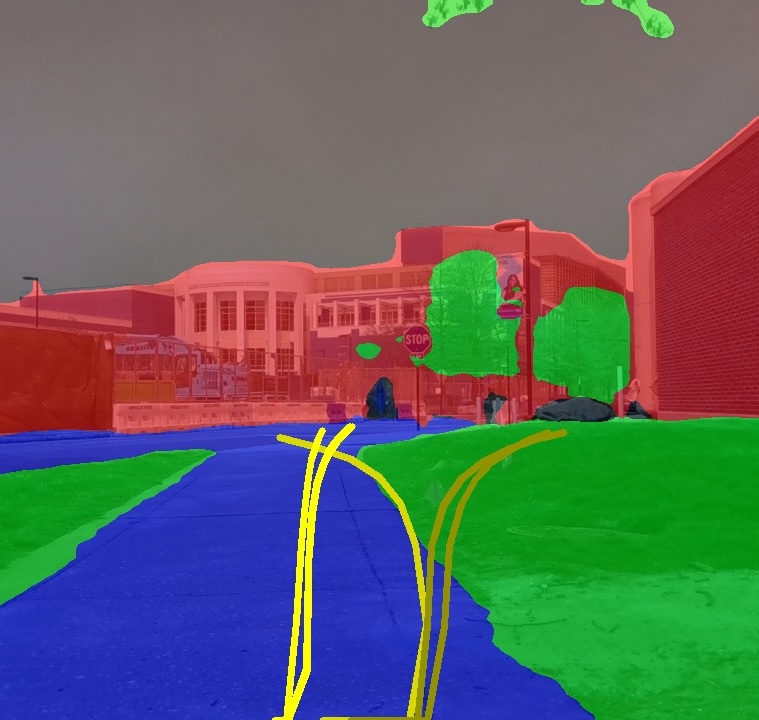}  (b)
   \end{minipage} 
   & 
   \begin{minipage}{.24\linewidth}   \centering

   \includegraphics[width=\linewidth,height=0.8\linewidth]{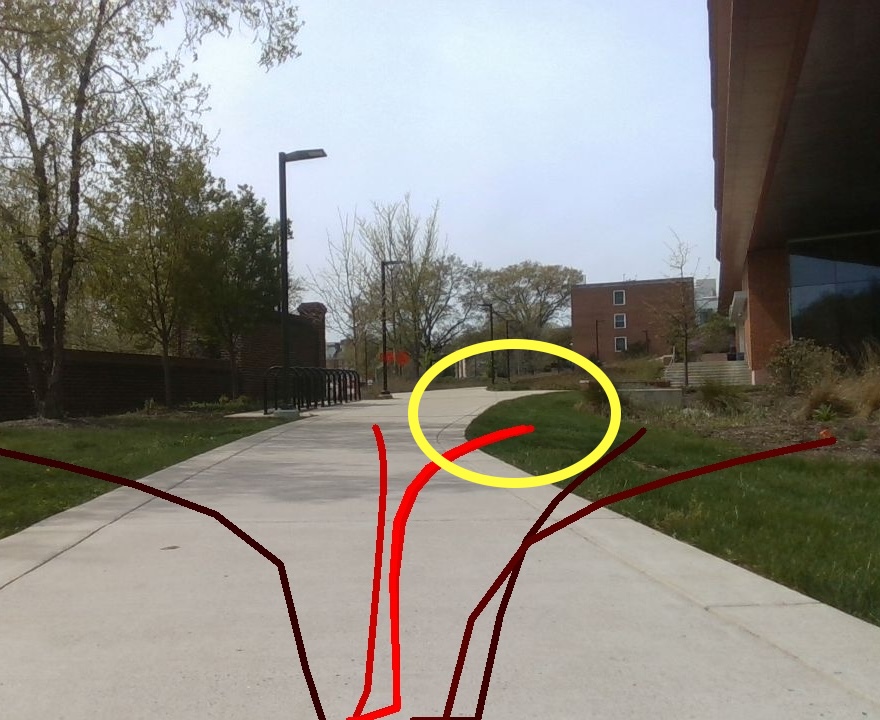} (c)
   \end{minipage}
   & 
   \begin{minipage}{.24\linewidth}    \centering
\includegraphics[width=\linewidth,height=0.8\linewidth]{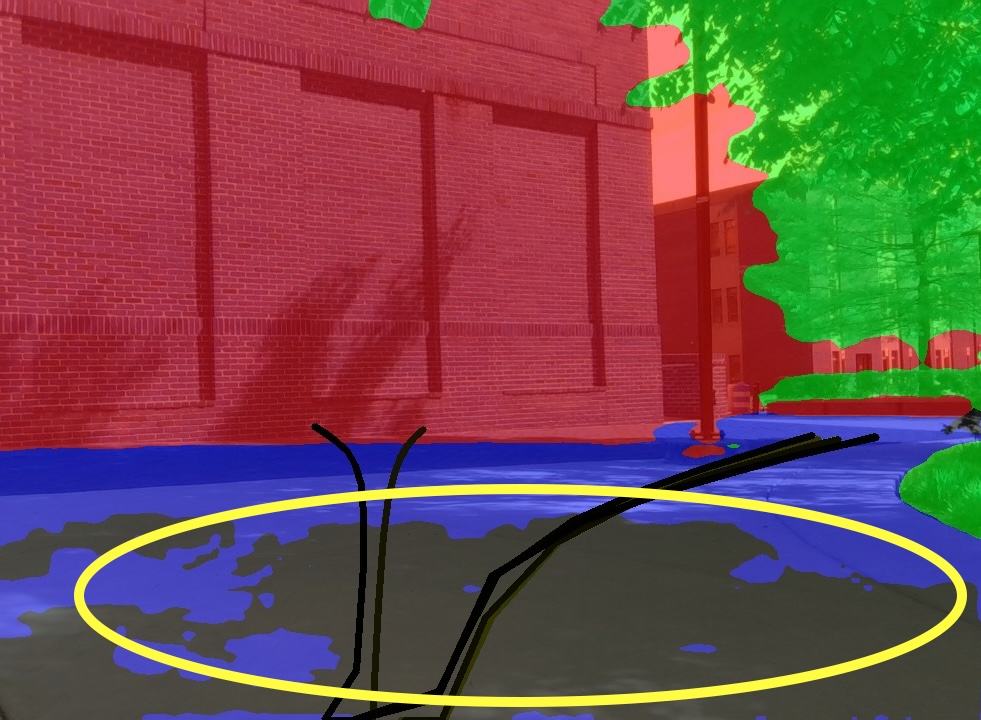} (d)\end{minipage}
  \end{tabular}
  % \vspace{-0.5em}
  \caption{ \small{\textbf{Geometric and Semantic Analysis:} Both geometric trajectory generation and semantic segmentation perform well in most cases, but there are also failure cases, as marked in yellow circles in (c) and (d). When the elevations of the lawn and sidewalk are similar, the geometric model struggles to perform accurately. Additionally, segmentation fails in out-of-distribution scenarios.}}
    \label{fig:challenge}
    % \vspace{-0.5em}
\end{figure}

\setlength{\tabcolsep}{4pt}

\begin{table}[htb]
% \vspace{-1.5em}
\centering
\scriptsize
\begin{tabularx}{\textwidth}{c|c|c|c|c}
\multirow{2}{*}{\textbf{Method}} & \multirow{2}{*}{\textbf{Modality}} & \multirow{2}{*}{\textbf{Traversability (\%) $\uparrow$}} & \textbf{Distance to} & \textbf{Inference} \\ 
 &  &  & \textbf{Target (\%) $\uparrow$} & \textbf{Time (s) $\downarrow$} \\ \hline
PIVOT & RGB + Language & 70 & 69 & 2.30 \\ \hline
ViNT & RGB & 57 & 62 & 0.69 \\\hline
NoMaD & RGB & 59 & 61 & 0.24 \\\hline
MTG~\cite{liang2024mtg} & Points & 61 & 64 & 0.01 \\ \hline
DTG~\cite{liang2024dtg} & Points & 67 & 66 & 0.13 \\ \hline
VL-TGS~\cite{song2024tgs} & Points, RGB, Language & 65 & 70 & 2.31 \\ \hline
\textbf{MOSU (ours)} & Points, RGB, Language & \textbf{77} & \textbf{73} & 2.30 \\
\end{tabularx}
\vspace{0.5em}
\caption{\textbf{Quantitative Evaluation:} Our approach achieves the best Traversability and comparable Distance to Target.}
\label{tab:gnd_quantitative}
\end{table}

% \begin{table}[htb]
% \vspace{-1.5em}
%     \centering
%     \begin{tabular}{c|c|c|c|c|c|c|c}
%       Methods         &   PIVOT & ViNT & NoMaD & MTG~\cite{liang2024mtg} & DTG~\cite{liang2024dtg} &  VL-TGS~\cite{song2024tgs} & MOSU (ours) \\ \hline
      
%       \multirow{2}{*}{Modality} & RGB & RGB & RGB & \multirow{2}{*}{Points} & \multirow{2}{*}{Points}  & Points, RGB,  & Points, RGB,\\
%                        &    Language  &  &    & Language & Language\\ \hline 
%       Traversability (\%) $\uparrow$    & 70 & x & x & 61 & 67 & 65 & 77     \\ \hline
%       Distance to Target ($\%$) $\uparrow$  & 69 & x & x & 64 & 66 & 70 & 73     \\ \hline
%       Inference Time (s) $\downarrow$  & 2.30 & 0.69 & 0.24 & 0.01 & 0.13 & 2.31 & 2.30
%     \end{tabular}
%     \caption{\textbf{Quantitative Evaluation}: Our approach achieves the best Traversability and comparable Distance to Target.
%     }
%     % \vspace{-1em}
%     \label{tab:gnd_quantitative}
% \end{table}

% \vspace{-0.5em}

\subsection{Experimental Insights}
% Main Experimental Insights (two pages)

As shown in Fig.~\ref{fig:challenge} (a), geometry-based trajectory generation~\cite{liang2024mtg} performs well in scenarios where geometric structures are easily detected. However, when the elevations of off-road areas and sidewalks are similar, it often fails, as shown in (c). In Fig.~\ref{fig:challenge} (a) and (b), lighter colors indicate higher scores in both geometric confidence and color-based traversability. As shown in (d), semantic segmentation also encounters out-of-distribution scenarios, where large ground areas cannot be properly segmented. Therefore, we use Equation~\ref{eq:trajectory} to integrate all components for optimal performance in traversability analysis. As shown in Table~\ref{tab:gnd_quantitative}, our approach achieves the best traversability among all methods, and the generated trajectories lead to the target closely.

As shown in Fig.~\ref{fig:gnd_qualitative}, beyond traversability analysis, VLMs also process labeled trajectories and analyze social cues from the images. In the first row, the VLM ranks the trajectories as [3, 0, 1, 2, 4], given the target at the right front. It suggests following trajectory 3 at normal speed, while other trajectories should be taken at a slower speed when encountering humans. The ranking for the second row is [4, 5, 0, 1, 2, 3] with normal speed. In the experiment, we observed that VLMs take a significant amount of time to process, as shown in Table~\ref{tab:gnd_quantitative}, but they demonstrate high accuracy in understanding social cues, particularly in detecting movement directions.

Besides social and traffic understanding, as shown in Tab.~\ref{tab:gnd_quantitative}, our approach achieves a comparable distance-to-target while attaining at least 10\% higher traversability than other methods.

\section{Conclusion}
We propose a system for long-range navigation that considers traversability, as well as social and traffic constraints. The system integrates routing, trajectory generation, and motion planning, leveraging the benefits of geometric, semantic, and language information to enhance scene understanding and trajectory generation. Compared with other state-of-the-art (SOTA) approaches, our method achieves a comparable distance-to-target and improves traversability by at least 10\%.

%\textbf{Limitations:} 
While the system demonstrates strong overall performance,  some limitations remain. It has difficulty in detecting small cliffs, such as the vertical surfaces of ramps. When the robot is moving on ramps, it is challenging to detect very low vertical surfaces.  Additionally, as with many learning-based approaches, the trajectory generation model lacks generalizability to out-of-distribution scenarios, which can lead to noisy trajectories and poor integration of geometric information. %\textbf{Future: } 
Future improvements may involve incorporating more robust geometric analysis methods, such as foundation geometric understanding models, to better evaluate geometric constraints.

% References (one page)
\bibliographystyle{styles/bibtex/splncs_srt}
\bibliography{references}

\end{document}